\documentclass[sigconf]{acmart}
\usepackage{makecell}
\AtBeginDocument{%
  }

\copyrightyear{2026}
\acmYear{2026}
\setcopyright{cc}
\setcctype{by}
\acmConference[MM '26]{Proceedings of the 34th ACM International Conference on Multimedia}{November 10--14, 2026}{Rio de Janeiro, Brazil}
\acmBooktitle{Proceedings of the 34th ACM International Conference on Multimedia (MM '26), November 10--14, 2026, Rio de Janeiro, Brazil}
\acmDOI{10.1145/3767308.3835072}
\acmISBN{979-8-4007-2213-4/2026/11}

\usepackage{multirow}

\begin{document}

\title{Do We Really Need Multimodal Emotion Language Models Larger Than 1B Parameters?}

\settopmatter{authorsperrow=3}

\author{Kaiwen Zheng}
\authornote{Equal contribution.}
\affiliation{%
  \institution{University of Glasgow}
  \country{United Kingdom}
}
\email{k.zheng.1@research.gla.ac.uk}

\author{Junchen Fu}
\authornotemark[1]
\affiliation{%
  \institution{University of Glasgow}
  \country{United Kingdom}
}
\email{j.fu.3@research.gla.ac.uk}

\author{Wenhao Deng}
\affiliation{%
  \institution{University of Glasgow}
  \country{United Kingdom}
}
\email{w.deng.1@research.gla.ac.uk}

\author{Hu Han}
\affiliation{%
  \institution{Institute of Computing Technology, Chinese Academy of Sciences}
  \country{China}
}
\email{hanhu@ict.ac.cn}

\author{Joemon M. Jose}
\affiliation{%
  \institution{University of Glasgow}
  \country{United Kingdom}
}
\email{Joemon.Jose@glasgow.ac.uk}

\author{Xuri Ge}
\authornote{Corresponding Author. Xuri Ge is also a member of the Qingdao Key Laboratory of Trustworthy Artificial Intelligence, Shandong University, China.}
\affiliation{%
  \institution{School of Artificial Intelligence, }
  \institution{Shandong University}
  \country{China}
}
\email{xuri.ge@sdu.edu.cn}

\renewcommand{\shortauthors}{Kaiwen Zheng et al.}

\begin{abstract}
Recent advances in multimodal large language models (MLLMs) have significantly improved the performance of multimodal emotion recognition (MER) and enabled interpretable description generation by jointly modeling video, audio, and language, etc. However, these performance improvements are often accompanied by an increase in model parameter size (e.g, at least 7B), which simultaneously incurs high computational costs and reduces inference efficiency, thereby hindering real-time deployment on resource-constrained platforms such as robots and mobile devices. This raises a fundamental question: do we really need the multimodal MER model larger than 1B parameters for high-quality MER? 

In this paper, we challenge the assumption that larger models are inherently necessary and proposes a lightweight MER framework (called \textbf{Light-MER}), which achieves better and faster multimodal sentiment understanding and recognition through knowledge distillation.
It can transfer knowledge from a strong, large-scale teacher model to a lightweight sub-billion-parameter student model, aiming to preserve rich multimodal emotion reasoning and recognition while substantially improving deployment efficiency.
Specifically, we introduce two new optimization strategies to enhance knowledge transfer: (1) a new optimal transport loss that combines Sliced Wasserstein Distance with hidden-state alignment, and (2) a new multi-reward optimization strategy based on GRPO that balances MER performance and efficiency, aimed at further enhancing the learning capabilities of student models. Extensive experiments on nine benchmark datasets demonstrate that Light-MER achieves state-of-the-art performance while significantly improving inference efficiency. This highlights the strong potential of small multimodal emotion language models for future research. Code is available at https://github.com/GAIR-Lab/Light-MER.
\end{abstract}



\maketitle
\begin{figure}[t]
    \centering
    \includegraphics[width=\linewidth]{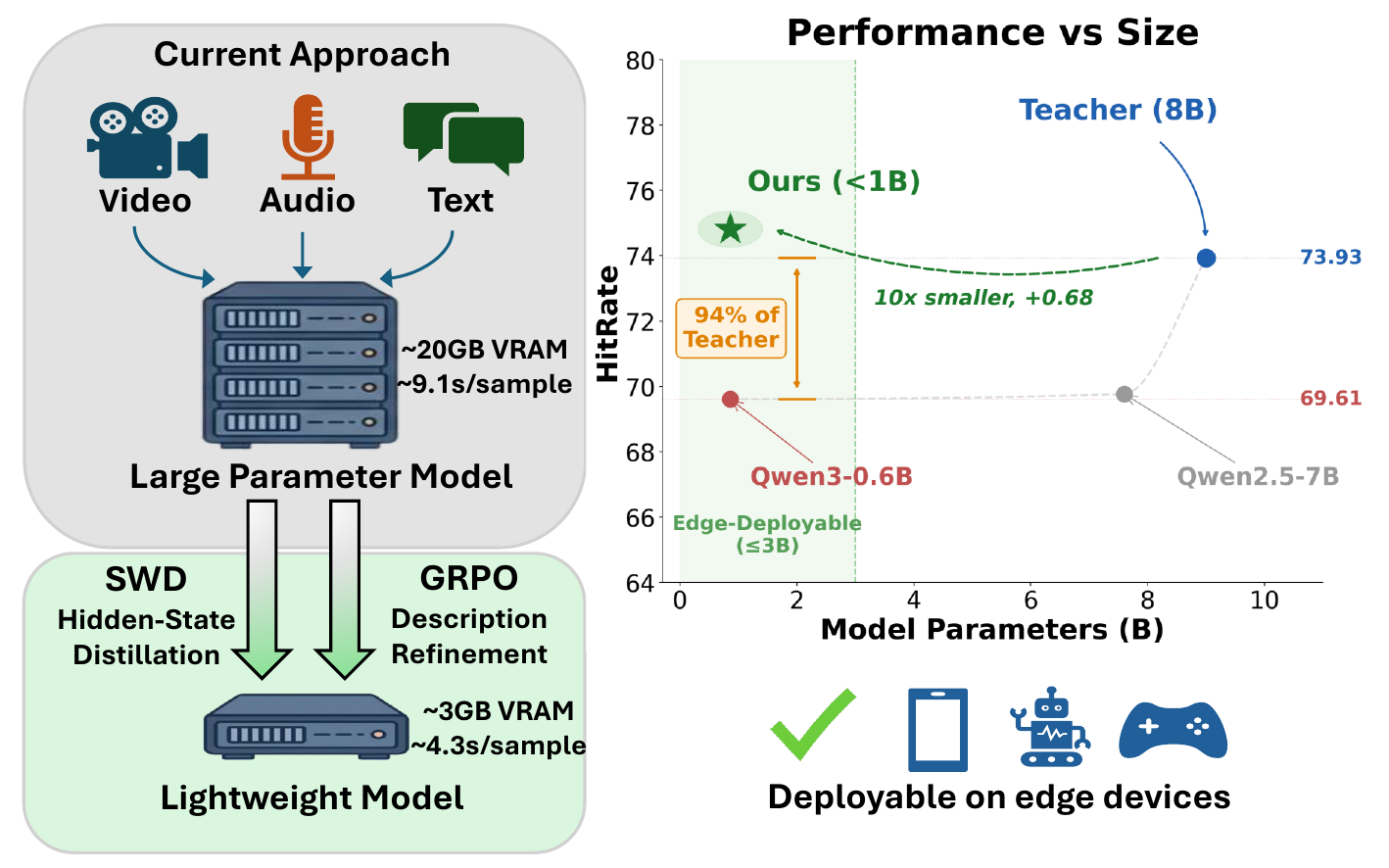}
    \vspace{-2em}
    \caption{Large multimodal emotion models achieve strong performance but are difficult to deploy at the edge. Our goal is to preserve most of the performance of an 8B teacher while moving the deployment model below 1B parameters.}
    \label{fig:motivation}
    \vspace{-1.5em}
\end{figure}
\section{Introduction}

Multimodal Emotion Recognition (MER) serves as a cornerstone for developing perceptive intelligent systems, demonstrating immense potential across domains such as social robotics \cite{MERRob}, healthcare assistance \cite{MERhealthcare}, and smart education\cite{gan2022iot}. Because affective expressions are inherently complex, linguistic cues alone are often insufficient for robust inference. Facial expressions, acoustic signals, scene dynamics, and textual context instead provide complementary evidence. A practical MER framework must therefore integrate visual, audio, and textual modalities while remaining computationally efficient under real-world resource constraints.

Existing MER methods can be broadly divided into discriminative and generative paradigms. Traditional MER \cite{TowardXuri} is mostly discriminative, where multimodal features are mapped to predefined emotion categories through task-specific classifiers. Although effective, these methods are limited to closed-set prediction and offer little interpretability. Recently, the rapid development of multimodal large language models (MLLMs) \cite{Yin_2024} has driven MER toward a more generative paradigm. Instead of predicting only a label, generative MER models \cite{zhao2026generative} can describe emotional states in natural language and provide richer evidence for their predictions. For example, AffectGPT \cite{affectgpt} shows that multimodal generators can capture more nuanced and compositional affective states than conventional classifiers. This shift is particularly important because human emotions are often subtle, ambiguous, and co-occurring.

Despite this progress, as illustrated in Figure \ref{fig:motivation}, current generative MER systems based on pre-trained MLLMs remain heavily reliant on increasingly large model scales—typically requiring at least 7B parameters \cite{EVLM2025, Emotion-LLaMA}—to achieve strong recognition and generation performance.
Although such scaling improves benchmark results, it also introduces substantial computational overhead, memory consumption, and inference latency, which severely limit deployment on robots, smartphones, and other edge devices. More fundamentally, generative MER requires the model to both integrate heterogeneous multimodal evidence and produce coherent emotion-aware descriptions, making efficient deployment substantially more challenging than classification-based MER. This naturally raises a key question: \textit{do we really need multimodal emotion language models larger than 1B parameters for high-quality MER?}

We argue that this is not necessary, provided that the multimodal reasoning abilities of large generative models can be effectively distilled into compact deployment models.
This motivates us to revisit MER from the perspective of efficient multimodal knowledge transfer. As illustrated in Figure~\ref{fig:motivation}, practical MER systems require a better balance among model size, accuracy, interpretability, and inference efficiency than current large-model solutions can provide. A natural way to achieve this goal is to first leverage a strong large-scale MER model as a source of rich multimodal knowledge, and then transfer such knowledge into a lightweight model suitable for deployment. In this way, as shown in Figure \ref{fig:motivation}, the large model serves as a teacher, while the compact model acts as a student.
 
The main challenge lies in how to preserve the teacher’s internal multimodal reasoning process during compression. Existing output-level distillation \cite{gu2026mini} based on KL divergence \cite{KL2014} mainly constrains the final token distribution, but does not preserve the latent reasoning structure that supports emotion understanding. Likewise, simple hidden-state regression losses such as MSE \cite{sun2019} ignore the distributional geometry of the teacher representation space. For MER, where subtle cues from voice, face, motion, and text must be integrated into coherent latent affective semantics, preserving such internal geometry is crucial for effective transfer. Therefore, a high-quality compact MER model requires not only model compression, but also a distillation framework that better preserves multimodal representation structure and generative capability.

To this end, we propose \textbf{Light-MER}, a knowledge distillation framework with multi-reward GRPO refinement for better and faster multimodal emotion understanding and recognition. Light-MER transfers knowledge from a strong large-scale MLLM to a lightweight sub-billion-parameter student model, aiming to retain rich multimodal emotion reasoning while substantially improving deployment efficiency. Specifically, our framework introduces two key optimization strategies. 
First, we define SWD-H as a hidden-state alignment objective that utilizes the Sliced Wasserstein Distance to minimize the distance between teacher and student latent distributions. Unlike point-wise alignment, SWD-H treats hidden representations as empirical distributions, facilitating an optimal transport-based mapping that is robust to high-dimensional manifold shifts.
Second, we further refine the distilled student with a new GRPO-based optimization strategy (called M-GRPO), designed with simultaneous constraints on multi-reward performance and efficiency for generative MER. 
We hypothesize that while hidden-state alignment (SWD-H) captures the teacher’s latent geometry, it does not inherently optimize for generative output quality. Therefore, we introduce M-GRPO as a critical refinement stage to align the model’s generated narratives with multi-faceted rewards, specifically focusing on the nuance of MER.

Our contributions are fourfold:
\begin{itemize}
    \item We revisit multimodal emotion recognition from an efficiency perspective and ask whether high-quality generative MER truly requires deployment models larger than 1B parameters.
    \item We propose Light-MER, a compact teacher-student framework that transfers multimodal emotion understanding and recognition ability from a large generative MLLM to a lightweight student model for deployment. 
    \item We introduce SWD-H, a hidden-state alignment objective based on Sliced Wasserstein Distance that explicitly preserves the geometry of multimodal latent structures, thereby improving the effectiveness of knowledge distillation.
    \item We propose M-GRPO, a new GRPO-based optimization strategy with simultaneous constraints on multi-reward performance and efficiency, to further improve generation quality and student learning for generative MER.
\end{itemize}
Extensive experiments on nine benchmark datasets demonstrate that Light-MER not only significantly improves inference efficiency, but also achieves state-of-the-art performance, showing that high-quality generative MER does not necessarily require deployment models larger than 1B parameters.

\section{Related Work}
\noindent \emph{Generative Multimodal Emotion Recognition.}
Early MER is mainly formulated as a discriminative classification problem, where multimodal features are fused and mapped to a predefined label space through task-specific classifiers.
Representative methods include multi-task learning with modality-specific encoders \cite{zheng-etal-2023-facial,zheng2026focal}, graph-based cross-modal reasoning \cite{Li_2023}, and attention-based fusion \cite{shi-huang-2023-multiemo}.
While effective in closed-label settings, these methods are limited in modeling ambiguity, co-occurrence, and fine-grained affective intensity.
Recent surveys summarize progress in datasets, fusion strategies, and model architectures \cite{biomimetics10070418, AV2024102218,zheng2025multimodal}.

Multimodal large language models (MLLMs) \cite{openai2024,yang2025,geminiteam2025,fu2025efficient,fu2026differentiable,fu2024exploring,he2025double,fu2026streamaware,deng2026recreclatentinterestsrecursive} have enabled generative MER, in which models describe emotions in natural language and support open-vocabulary reasoning. Emotion-LLaMA \cite{Emotion-LLaMA} unifies video, audio, and textual emotion understanding within a large language model, while AffectGPT \cite{affectgpt} combines multimodal encoders with a 7B language decoder and introduces MER-Caption+. A recent survey further highlights the role of MLLMs in emotion recognition and reasoning \cite{shou2025}. However, existing systems typically require deployment models with at least 7B parameters \cite{affectgpt,Emotion-LLaMA,su2023, zhang2026}, resulting in substantial memory and latency costs. Our work follows the generative paradigm while challenging the need for a large deployment model.

\noindent \emph{Knowledge Distillation and Optimal Transport Alignment.}
Knowledge distillation \cite{hinton2015} transfers capabilities from models to compact students. Most methods \cite{gu2026mini,agarwal2024}, including DistilBERT \cite{sanh2020}, use softened cross-entropy or KL divergence over output distributions. Logit-level objectives are less suitable for generative MER, where multimodal reasoning must be preserved and peaked teacher distributions provide little supervision beyond the top token.

Richer objectives include layer-wise hidden-state alignment in TinyBERT \cite{jiao2020}, reverse-KL output distillation in MiniLLM \cite{gu2026mini}, and recent analyses supporting hidden-state alignment \cite{dasgupta2025improving}. However, pointwise MSE and cosine objectives do not preserve representation geometry explicitly, while CKA-based alignment \cite{zhou2024} addresses it only partially. This is particularly limiting when hidden states organize cross-modal evidence before decoding.

Optimal transport (OT) offers a principled approach to geometry-aware alignment \cite{cui2024,lv2024}. Compared with KL-based objectives, Wasserstein distances are more robust under partial distribution mismatch. However, existing OT variants such as Sinkhorn OT \cite{NIPS2013_af21d0c9}, position-aware OT \cite{PAOT}, and unbalanced OT \cite{UnbalancedOT} either require iterative optimization or introduce additional coupling and balancing terms, increasing computational cost in repeated hidden-state matching. We therefore adopt Sliced Wasserstein Distance (SWD) \cite{kolouri2019}, which enables efficient OT approximation via random one-dimensional projections and sorting. SWD retains the geometric benefits of OT with much lower overhead and has shown effectiveness in recent compression settings \cite{quetu2025}. Accordingly, we use SWD to align teacher and student hidden-state distributions for generative MER.

\noindent \emph{Reinforcement Learning for Post-distillation Refinement.}
Reinforcement learning is widely used to improve language generation. RLHF \cite{ouyang2022} and PPO \cite{schulman2017} are effective but typically require a value model or critic. GRPO \cite{shao2024} removes this requirement by estimating group-relative advantages from multiple responses per prompt, simplifying training. It has been adopted by reasoning-oriented models such as DeepSeekMath \cite{shao2024} and DeepSeek-R1 \cite{Guo_2025}.

We apply post-distillation GRPO with rewards for emotion accuracy, format, conciseness, suppression of unnecessary reasoning, clean stopping, and label-count control, complementing representation-level distillation to improve quality and efficiency.

\begin{figure*}[t!]
    \centering
    \includegraphics[width=\textwidth]{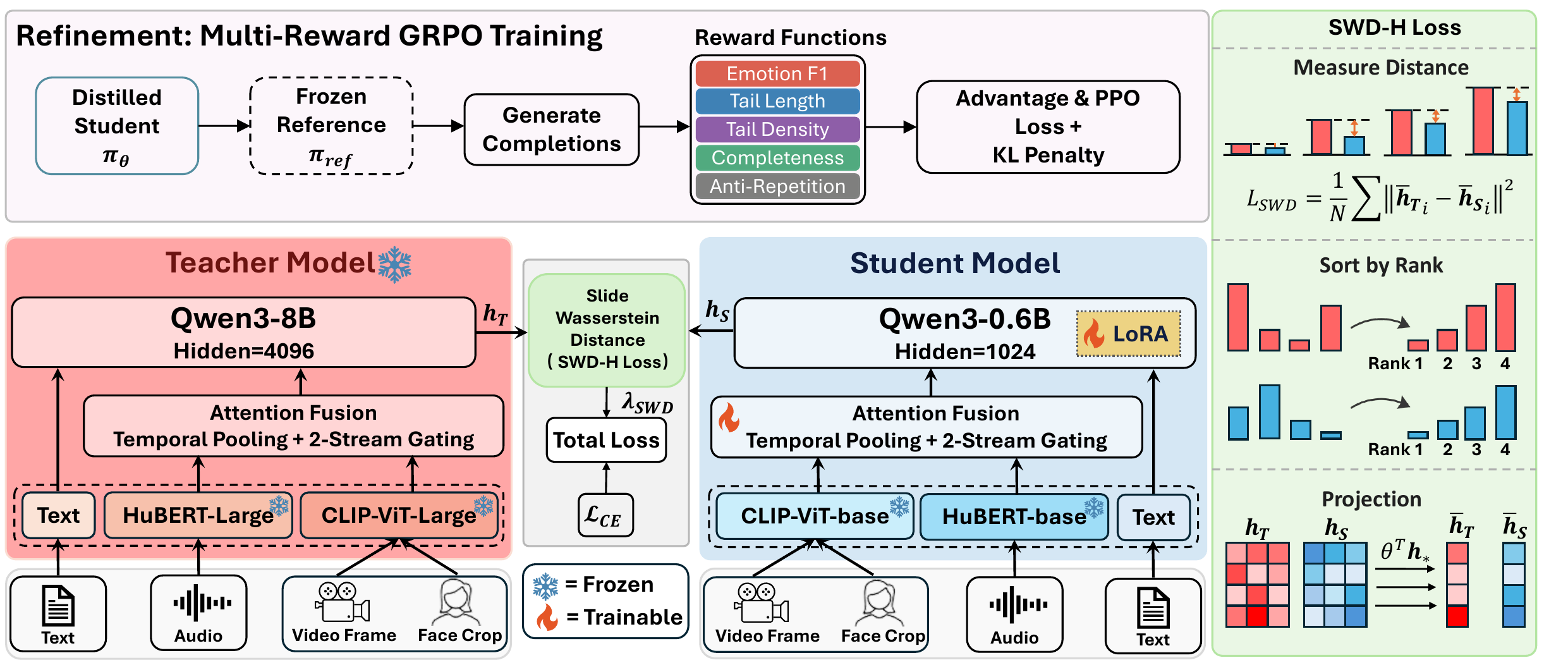}
    \vspace{-2em}
    \caption{An overview of our model. A frozen 8B teacher supervises a sub-1B student through SWD hidden-state alignment (SWD-H), and the distilled student is further refined with multi-reward GRPO for concise emotion description generation.}
    \label{fig:model}
    \vspace{-1em}
\end{figure*}

\section{Method}

\subsection{Problem Formulation}

Given a multimodal sample $\mathcal{X}=\{x^{\mathrm{frm}}/x^{\mathrm{face}},x^{\mathrm{aud}},x^{\mathrm{sub}}\}$ and an instruction $q$, the model generates an emotion description $\mathcal{Y}=\{y_1,\ldots,y_L\}$. The visual stream supports either full video frames $x^{\mathrm{frm}}$ or face-cropped clips $x^{\mathrm{face}}$; as facial regions carry the most salient affective cues, we adopt face crops in all experiments. We consider a teacher model $\mathcal{T}$ and a student model $\mathcal{S}$. The teacher is a high-capacity multimodal emotion language model used only during training, while the student is the deployment model. Our goal is to train $\mathcal{S}$ to (i) preserve the teacher's multimodal emotion understanding ability through hidden-state distillation, and (ii) produce shorter and cleaner descriptions through policy optimization.
The standard autoregressive objective is
\begin{equation}
\mathcal{L}_{\mathrm{CE}}
=
-\frac{1}{L}\sum_{l=1}^{L}\log p_{\theta}(y_l\mid x,q,y_{[1:l-1]}),
\end{equation}
where $\theta$ denotes the student parameters.

Building a distillation framework for this objective requires two design choices: (i) at which representation level should teacher and student be aligned (output logits, intermediate hidden states, or cross-layer relations), and (ii) which divergence measure should drive the alignment. We address (i) through an empirical analysis of hidden-state and output distributions in Section~\ref{subsec:observation}, and (ii) through the SWD formulation in Section~\ref{subsec:swd}. The student is further refined with GRPO-based multi-reward policy optimization in Section~\ref{M-GRPO}.

\subsection{Model Architecture}

Figure~\ref{fig:model} summarizes the framework of Light-MER.
The teacher consists of Qwen3-8B \cite{yang2025} as the language decoder,
CLIP-ViT-Large-Patch14 \cite{radford2021} as the visual encoder, and
HuBERT-Large \cite{hsu2021hubert} as the acoustic encoder.
It is pre-trained on the MER-Caption+ \cite{affectgpt} corpus and frozen
during student distillation.
The student replaces the teacher with a deployment-scale architecture:
Qwen3-0.6B \cite{yang2025} as the language decoder,
CLIP-ViT-Base-Patch16 \cite{radford2021} as the visual encoder, and
HuBERT-Base \cite{hsu2021hubert} as the acoustic encoder.
The LLM branch is updated through LoRA while the multimodal projectors
and fusion layers are trained jointly with the student decoder.
The central design principle is that the teacher should remain large
enough to provide rich multimodal hidden-state supervision, while the
student should remain small enough for practical deployment.

\paragraph{Multimodal Fusion.}
Each modality stream $m\!\in\!\{\mathrm{face},\mathrm{aud}\}$ is first
encoded by a corresponding frozen encoder into a temporal sequence
$\mathbf{Z}^{(m)}\!=\!\{\mathbf{z}_1^{(m)},\ldots,\mathbf{z}_{T_m}^{(m)}\}$,
then compressed into a single vector via learned temporal pooling
$\bar{\mathbf{z}}^{(m)}\!=\!\sum_t a_t^{(m)}\mathbf{z}_t^{(m)}$
and projected into the LLM input space by a linear mapping
$\mathbf{h}^{(m)}\!=\!\mathbf{W}_m\bar{\mathbf{z}}^{(m)}$.
For multimodal fusion, the two encoder sequences are independently
mean-pooled and mapped to a common feature space:
\begin{equation}
\mathbf{u}^{(m)}
=
\mathbf{P}_m
\left(
\frac{1}{T_m}\sum_t\mathbf{z}_t^{(m)}
\right).
\end{equation}
The two streams are then merged using learned modality weights:
\begin{equation}
\mathbf{h}^{(\mathrm{mm})}
=
\mathbf{W}_{\mathrm{mm}}
\left(
\beta_{\mathrm{face}}\mathbf{u}^{(\mathrm{face})}
+
\beta_{\mathrm{aud}}\mathbf{u}^{(\mathrm{aud})}
\right),
\end{equation}
\begin{equation}
[\beta_{\mathrm{face}},\beta_{\mathrm{aud}}]
=
\mathbf{W}_2\mathbf{W}_1
[\mathbf{u}^{(\mathrm{face})}:\mathbf{u}^{(\mathrm{aud})}],
\end{equation}
where $[:]$ indicates concatenation.
The LLM receives the fused multimodal token $\mathbf{h}^{(\mathrm{mm})}$
together with separate face tokens, audio tokens, the text subtitle,
and the instruction prompt, and decodes via the Qwen3-0.6B student
with LoRA adaptation.
All reported results use this gating-based fusion with face inputs.

\subsection{Empirical Observation: Hidden States vs.\ Output Distributions}
\label{subsec:observation}

\begin{figure*}[t]
    \centering
    \includegraphics[width=0.9\textwidth]{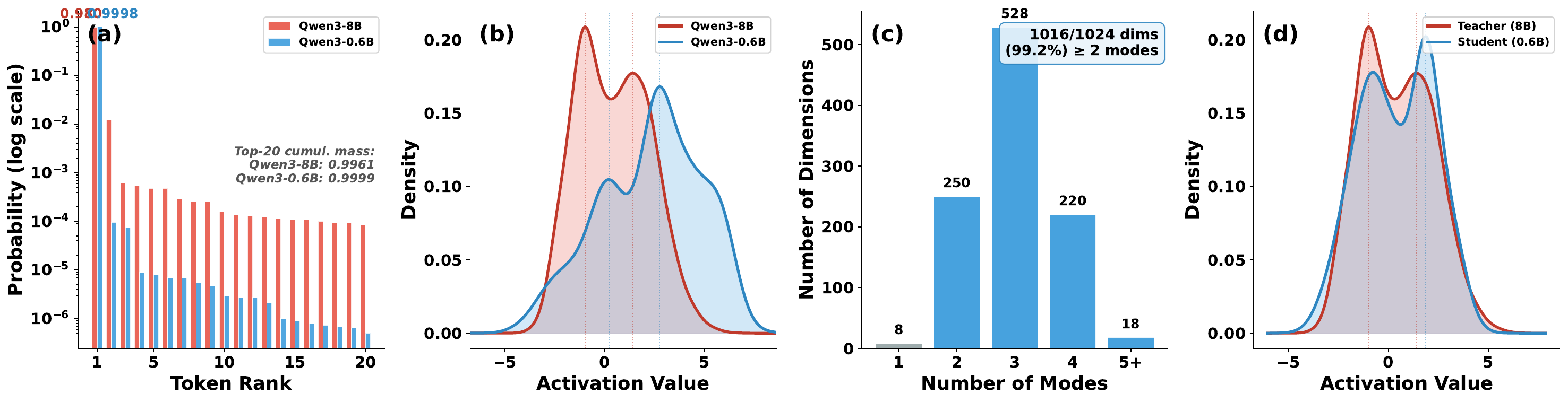}
    \vspace{-1.0em}
    \caption{Hidden-state and output-probability analysis aggregated across subsets evaluation suites.
    (a) Top-20 token probabilities.
    (b) A representative hidden-state dimension before distillation. Teacher and student are both multi-modal with different mode locations.
    (c) Mode-count histogram across all 1024 dimensions. 99.2\% of dimensions contain two or more modes.
    (d) The same dimension as (b) after SWD-H distillation. Teacher and student modes are more aligned.}
    \label{fig:hidden_logit_analysis}
    \vspace{-0.6em}
\end{figure*}

Before distillation begins, we compare the 8B and 0.6B models that have been independently trained on the MER task (Figure~\ref{fig:hidden_logit_analysis}). We aggregate statistics across subsets sampled from the nine-benchmark suite and examine output probabilities and last-layer hidden states.

Output probabilities are extremely peaked for both models (Figure~\ref{fig:hidden_logit_analysis}a). The 8B model places 0.980 of its probability on the top-1 token. Tokens ranked 2--20 collectively carry only 0.016 of the 8B model's mass. When the teacher's distribution is this peaked, KL-based distillation behaves similarly to hard-label cross-entropy training (the gradient of KL divergence at each token is weighted by the teacher's probability, so the top-1 token dominates), leaving little room for the student to learn from the teacher's output distribution beyond the top prediction.

Across the same aggregated data, hidden-state activations exhibit multi-modal structure in 1016 of 1024 dimensions, or 99.2\% (Figure~\ref{fig:hidden_logit_analysis}c). A representative sample (Figure~\ref{fig:hidden_logit_analysis}b) illustrates the pattern: the 8B and 0.6B models place modes at different positions and heights. Despite producing nearly identical outputs, the two models organize their internal representations differently. OT-based objectives can account for the distance between these modes, while pointwise losses such as MSE only penalize per-position differences without considering the overall distribution shape. The SWD-H formulation below builds on this observation. A visual comparison before and after SWD-H training is shown in Figure~\ref{fig:hidden_logit_analysis}(b,\,d) and discussed in Section~\ref{sec:ablation}.

\subsection{Sliced Wasserstein Distillation on Hidden States (SWD-H)}
\label{subsec:swd}

Our distillation target is the last-layer hidden state of the LLM rather than the output logits. Let $\mathbf{H}^{T}\in\mathbb{R}^{L\times4096}$ and $\mathbf{H}^{S}\in\mathbb{R}^{L\times1024}$ denote the teacher and student hidden states at the supervised decoding positions. To compare them in the same space, we use an SWD projector composed of a frozen orthogonally initialized teacher projection, an identity-initialized student projection, and LayerNorm:
\begin{equation}
\tilde{\mathbf{H}}^{T}=\mathbf{H}^{T}\mathbf{W}_{T},
\qquad
\tilde{\mathbf{H}}^{S}=\mathrm{LN}(\mathbf{H}^{S}\mathbf{W}_{S}),
\end{equation}
where $\mathbf{W}_T\in\mathbb{R}^{4096\times1024}$ is frozen and $\mathbf{W}_S\in\mathbb{R}^{1024\times1024}$ is identity-initialized. Following standard autoregressive training, the cross-entropy loss is computed only at positions whose next token belongs to the answer sequence. We apply the same shifted mask as cross-entropy to the hidden-state distillation: for each position $i$ in the sequence, SWD alignment is active if and only if $y_{i+1} \neq -100$ (Ignore Index), where $y$ denotes the label sequence.

We then sample $R=200$ random unit directions $\{\boldsymbol{\theta}_r\}_{r=1}^{R}$ in $\mathbb{R}^{1024}$ and compute one-dimensional projections for teacher and student. The SWD-H loss is
\begin{equation}
\mathcal{L}_{\mathrm{SWD}}
=
\frac{1}{Rn}
\sum_{r=1}^{R}
\left\|
\mathrm{sort}\!\left(\tilde{\mathbf{H}}^{T}_{M}\boldsymbol{\theta}_r\right)
-
\mathrm{sort}\!\left(\tilde{\mathbf{H}}^{S}_{M}\boldsymbol{\theta}_r\right)
\right\|_1,
\end{equation}
where $M$ denotes the answer-token mask. In one dimension, the optimal transport plan between two empirical distributions of equal size $n$ is the quantile coupling: sorting both sets of values and matching them index-by-index is provably optimal \cite{peyre2020}. The $p$-Wasserstein distance therefore reduces to
\begin{equation}
W_p^p(\alpha,\beta)
=
\frac{1}{n}\sum_{i=1}^{n}\bigl|\alpha_{(i)}-\beta_{(i)}\bigr|^{p},
\end{equation}
where $\alpha_{(1)}\!\leq\!\cdots\!\leq\!\alpha_{(n)}$ and $\beta_{(1)}\!\leq\!\cdots\!\leq\!\beta_{(n)}$ are the sorted projections ($p{=}1$ in our case). This closed-form solution gives $\mathcal{L}_{\mathrm{SWD}}$ an $O(Rn\log n)$ complexity dominated by sorting, compared with $O(Kn^{2})$ for $K$-iteration Sinkhorn, and eliminates the need for entropic regularization $\epsilon$.

The total distillation objective is
\begin{equation}
\mathcal{L}
=
\mathcal{L}_{\mathrm{CE}}
+
\alpha_t\,\mathcal{L}_{\mathrm{SWD}},
\qquad
\alpha_t
=
\lambda_{\mathrm{SWD}}
\min\!\left(\frac{t}{5000},1\right),
\end{equation}
where $\lambda_{\mathrm{SWD}}=1.0$ and the SWD-H weight $\alpha_t$ is linearly warmed up over the first 5000 steps. Across eight distillation variants evaluated in this study, SWD-H is the best overall strategy.

\subsection{Multi-Reward GRPO-based (M-GRPO) Description Refinement} \label{M-GRPO}

Distillation preserves the teacher's emotion reasoning ability, but it also transfers the teacher's verbose generation pattern, a common issue in large-scale affective models that produce over-elaborate emotional descriptions.
In our setting, the model output consists of a canonical prefix (e.g., ``The character's emotional state is \ldots'') followed by one or more open-set emotion labels. We apply GRPO to the complete generated response, including the canonical prefix. Since emotion F1 alone cannot control output format, verbosity, unnecessary reasoning, stopping behavior, or label count, we use a composite reward with complementary components.
We define the total reward as
\begin{equation}
r
=
2.0\,r_{\mathrm{F1}}
+
1.0\,r_{\mathrm{fmt}}
+
0.3\,r_{\mathrm{brev}}
+
0.5\,r_{\mathrm{noreason}}
+
0.3\,r_{\mathrm{stop}}
+
0.5\,r_{\mathrm{count}},
\end{equation}
where $r_{\mathrm{F1}}$ measures the F1 overlap between predicted labels and ground-truth labels, $r_{\mathrm{fmt}}$ rewards the canonical emotion-state format, $r_{\mathrm{brev}}$ encourages concise outputs, $r_{\mathrm{noreason}}$ penalizes unnecessary reasoning phrases, $r_{\mathrm{stop}}$ rewards clean single-sentence stopping, and $r_{\mathrm{count}}$ encourages a reasonable number of emotion labels.

\begin{table*}[t]
    \centering
    \small
    \setlength{\tabcolsep}{4pt}
    \renewcommand{\arraystretch}{0.9}
    \caption{Main evaluation results following the layout of AffectGPT Table~2. Metrics are reported for each dataset: HIT for basic emotion, WAF for sentiment, and $F_s$ for OV-MERD+. AffectGPT baselines are reproduced using the official open-source code.}
    \vspace{-1em}
    \label{tab:main_results}
    \resizebox{\textwidth}{!}{
    \begin{tabular}{lr|ccc|cccc|cccc|c|c}
        \toprule
        \multirow{2}{*}{Model} & \multirow{2}{*}{LLM Param} & \multicolumn{3}{c|}{Modality} & \multicolumn{4}{c|}{Basic Emotion} & \multicolumn{4}{c|}{Sentiment} & \multicolumn{1}{c|}{Fine-grained} & \multirow{2}{*}{Mean} \\
        & & A & V & T & MER2023 & MER2024 & MELD & IEMOCAP & MOSI & MOSEI & SIMS & SIMS v2 & OV-MERD+ & \\
        \midrule
        \multicolumn{15}{l}{\textbf{Audio+Text MLLMs}} \\
        OneLLM \cite{OneLLM}       & 7B & $\checkmark$ & $\times$     & $\checkmark$ & 25.52 & 17.21 & 28.32 & 33.44 & 64.01 & 54.09 & 63.39 & 61.98 & 22.25 & 41.14 \\
        SECap \cite{SECap}        & 7B & $\checkmark$ & $\times$     & $\checkmark$ & 40.95 & 52.46 & 25.56 & 36.92 & 55.76 & 54.18 & 59.51 & 57.41 & 36.97 & 46.64 \\
        PandaGPT \cite{su2023}     & 7B & $\checkmark$ & $\times$     & $\checkmark$ & 33.57 & 39.04 & 31.91 & 36.55 & 66.06 & 61.33 & 62.93 & 58.88 & 31.33 & 46.84 \\
        Qwen-Audio \cite{Qwen-Audio}  & 7B & $\checkmark$ & $\times$     & $\checkmark$ & 41.85 & 31.61 & 49.09 & 35.47 & 70.09 & 46.90 & 70.73 & 65.26 & 32.36 & 49.26 \\
        SALMONN \cite{SALMONN}     & 13B & $\checkmark$ & $\times$     & $\checkmark$ & 55.53 & 45.38 & 45.62 & \underline{46.84} & \underline{81.00} & 67.03 & 68.69 & 65.93 & 45.00 & 57.89 \\
        AffectGPT(Qwen3-8B) & 8B & $\checkmark$ & $\times$ & $\checkmark$ & \underline{59.28} & \underline{56.31} & \textbf{54.03} & 46.69 & \textbf{81.73} & \underline{78.56} & \underline{77.77} & \underline{78.85} & \underline{52.39} & \underline{65.07} \\
        \textbf{Light-MER(Ours)}  & \textbf{0.6B} & $\checkmark$ & $\times$ & $\checkmark$ & \textbf{59.33} & \textbf{58.30} & \underline{53.10} & \textbf{52.18} & 80.48 & \textbf{79.95} & \textbf{80.74} & \textbf{79.80} & \textbf{53.75} & \textbf{66.40} \\
        \midrule
        \multicolumn{15}{l}{\textbf{Video+Text MLLMs}} \\
        Otter \cite{Otter}           & 9B & $\times$     & $\checkmark$ & $\checkmark$ & 16.41 & 14.65 & 22.57 & 29.08 & 52.89 & 50.44 & 57.56 & 53.12 & 16.63 & 34.82 \\
        Video-LLaVA  \cite{video-llava}    & 7B & $\times$     & $\checkmark$ & $\checkmark$ & 36.93 & 30.25 & 30.73 & 38.95 & 56.37 & 61.64 & 53.28 & 57.45 & 34.00 & 44.40 \\
        PandaGPT \cite{su2023}        & 7B & $\times$     & $\checkmark$ & $\checkmark$ & 39.13 & 47.16 & 38.33 & 47.21 & 58.50 & 64.25 & 62.07 & 65.25 & 35.07 & 50.77 \\
        Video-ChatGPT \cite{video-chatgpt}   & 7B & $\times$     & $\checkmark$ & $\checkmark$ & 44.86 & 46.80 & 37.33 & \textbf{56.83} & 54.42 & 63.12 & 64.82 & 65.80 & 39.80 & 52.64 \\
        LLaMA-VID  \cite{llamavid}      & 7B & $\times$     & $\checkmark$ & $\checkmark$ & 50.72 & 57.60 & 42.75 & 46.02 & 61.78 & 63.89 & 69.35 & 67.48 & 45.01 & 56.07 \\
        VideoChat  \cite{videochat}      & 7B & $\times$     & $\checkmark$ & $\checkmark$ & 48.73 & 57.30 & 41.11 & 48.38 & 65.13 & 63.61 & 69.52 & 72.14 & 44.52 & 56.71 \\
        Chat-UniVi \cite{chatunivi}      & 7B & $\times$     & $\checkmark$ & $\checkmark$ & 57.62 & \textbf{65.67} & 45.61 & 52.37 & 54.53 & 63.18 & 68.15 & 66.36 & 48.00 & 57.94 \\
        mPLUG-Owl  \cite{mPLUG}      & 7B & $\times$     & $\checkmark$ & $\checkmark$ & 56.86 & \underline{59.89} & 49.11 & \underline{55.54} & 72.40 & 72.91 & 72.13 & 75.00 & 48.18 & 62.45 \\
        AffectGPT(Qwen3-8B) & 8B & $\times$ & $\checkmark$ & $\checkmark$ & \underline{58.70} & 55.28 & \underline{50.51} & 48.59 & \textbf{83.11} & \underline{75.50} & \underline{78.15} & \underline{80.49} & \underline{49.59} & \underline{64.44}\\
        \textbf{Light-MER(Ours)}  & \textbf{0.6B} & $\times$ & $\checkmark$ & $\checkmark$ & 58.77 & 56.31 & \textbf{51.51} & 47.38 & \underline{82.39} & \textbf{79.36} & \textbf{82.80} & \textbf{81.32} & \textbf{51.51} & \textbf{65.71} \\
        \midrule
        \multicolumn{15}{l}{\textbf{Audio+Video+Text MLLMs}} \\
        PandaGPT  \cite{su2023}       & 7B & $\checkmark$ & $\checkmark$ & $\checkmark$ & 40.21 & 51.89 & 37.88 & 44.04 & 61.92 & 67.61 & 68.38 & 67.23 & 37.12 & 52.92 \\
        Emotion-LLaMA \cite{Emotion-LLaMA}   & 7B & $\checkmark$ & $\checkmark$ & $\checkmark$ & 59.38 & 73.62 & 46.76 & 55.47 & 66.13 & 67.66 & 78.32 & 77.23 & 52.97 & 64.17 \\
        AffectGPT \cite{affectgpt} & 7B & $\checkmark$ & $\checkmark$ & $\checkmark$ & \textbf 69.74 & 67.08 & \underline{56.05} & 53.51 & \textbf{83.49} & \underline{78.06} & 82.79 & 80.15 & 57.03 & 69.77 \\
        AffectGPT(Qwen3-8B)& 8B & $\checkmark$ & $\checkmark$ & $\checkmark$ & \textbf{76.72} & \underline{79.68} & 55.99 & \underline{59.32} & 78.46 & \textbf{78.48} & \underline{88.30} & \underline{86.94} & \underline{61.47} & \underline{73.93} \\
        \textbf{Light-MER(Ours)}  & \textbf{0.6B} & $\checkmark$ & $\checkmark$ & $\checkmark$ & \underline{75.66} & \textbf{80.04} & \textbf{57.02} & \textbf{61.46} & \underline{80.04} & 77.23 & \textbf{89.05} & \textbf{87.74} & \textbf{63.28} & \textbf{74.61} \\
        \bottomrule
    \end{tabular}}
    \vspace{-0.5em}
\end{table*}

\begin{table*}[t]
    \centering
    \tiny
    \caption{Full-pipeline inference profiling.}
    \vspace{-1em}
    \fontsize{8.5}{8.5}\selectfont
    \renewcommand\tabcolsep{5pt}  
    \label{tab:inference-profile}
    \begin{tabular}{l c c c c cc cc}
    \toprule
    & & & & & \multicolumn{2}{c}{Direct} & \multicolumn{2}{c}{Descriptive} \\
    \cmidrule(lr){6-7} \cmidrule(lr){8-9}
    Model
      & \makecell{Total\\Params}
      & \makecell{FLOPs\\(G)}
      & \makecell{Peak\\Mem.}
      & Compress.
      & \makecell{Time\\(s/samp.)}
      & \makecell{Words\\/samp.}
      & \makecell{Time\\(s/samp.)}
      & \makecell{Words\\/samp.} \\
    \midrule
    Teacher (Qwen3-8B)         & 9.00B   & 10902.6 & 20.04\,GB & --           & 0.901 & 9.5   & 6.138 & 104.4 \\
    Student SWD-H (Qwen3-0.6B)   & 854.93M &   988.8 &  2.54\,GB & $11.0\times$ & 0.561 & 8.5   & 4.621 & 110.5 \\
     Student M-GRPO (Qwen3-0.6B)  & 854.93M &   988.8 &  2.54\,GB & $11.0\times$ & 0.523 & 7.9   & 3.105 & 70.8  \\
    \bottomrule
    \end{tabular}%
\end{table*}

For each training sample, we generate $G=4$ completions and compute group-relative advantages. This choice strikes a reasonable balance between group diversity and computational overhead.
\begin{equation}
A_i =
\frac{
r_i-\frac{1}{G}\sum_{j=1}^{G}r_j
}{
\sigma_G+10^{-8}
},
\end{equation}
where $\sigma_G$ is the sample standard deviation of the group rewards. When $\sigma_G<10^{-8}$, the advantages are set to zero and the group is skipped.

We then optimize a PPO-style clipped surrogate objective with an approximate KL penalty to the frozen distilled reference model:
\begin{equation}
\mathcal{L}_{\mathrm{GRPO}}
=
-\mathbb{E}_i
\left[
\min\!\left(
\rho_i A_i,
\mathrm{clip}(\rho_i,1-\epsilon,1+\epsilon)A_i
\right)
\right]
+
\beta\,\mathbb{E}_i[\mathrm{KL}_i],
\end{equation}
where $\epsilon=0.10$ and $\beta=1.0$. For numerical stability, we first clamp the sequence-level log-ratio:
\begin{equation}
\Delta_i=
\mathrm{clip}\!\left(
\log \pi_{\theta}(y_i\mid x)
-
\log \pi_{\mathrm{ref}}(y_i\mid x),
-5,5
\right),
\end{equation}
and define
\begin{equation}
\rho_i=\exp(\Delta_i),
\end{equation}
\begin{equation}
\mathrm{KL}_i
=
\rho_i-1-\Delta_i
=
\exp(\Delta_i)-1-\Delta_i.
\end{equation}
This stage encourages concise, well-formatted, and emotion-faithful outputs, making the distilled student efficient at inference time.

\section{Experimental Setup}

\subsection{Datasets}

We evaluate on nine benchmarks spanning three task families: basic emotion recognition (MER2023 \cite{MER2023}, MER2024 \cite{MER2024}, MELD \cite{MELD}, and IEMOCAP \cite{IEMOCAP}), sentiment analysis (CMU-MOSI \cite{CMU-MOSI}, CMU-MOSEI \cite{CMU-MOSEI}, CH-SIMS \cite{CH-SIMS}, and CH-SIMS v2 \cite{CH-SIMv2}), and fine-grained open-vocabulary MER (OV-MERD+ \cite{OV-MERD+}). This coverage is intentionally broad: a compact deployment model is only useful if it remains robust across fixed-label and open-set settings, sentiment and emotion tasks, and both Chinese and English data.

\subsection{Implementation Details}

The teacher uses Qwen3-8B, CLIP-ViT-Large-Patch14, and HuBERT-Large, and is trained for 60 epochs on MER-Caption+. The student uses Qwen3-0.6B, CLIP-ViT-Base-Patch16, and HuBERT-Base.

\subsection{Evaluation Protocol}

For basic emotion recognition, we use HIT rate as the primary metric: a prediction is counted as correct if the ground-truth basic label appears in the predicted label set after emotion-wheel grouping. We additionally report \texttt{mscore} for these datasets. For sentiment analysis, we report weighted average F-score (WAF) and accuracy (ACC), using WAF as the primary metric because of label imbalance; For fine-grained open-vocabulary MER, we follow the set-level formulation and report $\mathrm{Precision}_s$, $\mathrm{Recall}_s$, and $F_s$, averaged over five emotion wheels; $F_s$ is the primary metric. The dataset-wise mean reported in the main table averages the primary metric of each benchmark.

\section{Experiments}

We organize the evaluation around five Research Questions.

\noindent \textbf{RQ1 (Effectiveness):} Can a sub-1B model match or surpass an 8B teacher across diverse MER benchmarks?

\noindent \textbf{RQ2 (Component Ablation):} What does each component (SWD-H, M-GRPO) contribute to the final performance?

\noindent \textbf{RQ3 (Distillation Design):} Among OT-based distillation variants, does SWD-H perform highest?

\noindent \textbf{RQ4 (Generation Refinement):} Does M-GRPO improve generation quality while maintaining recognition accuracy?

\noindent \textbf{RQ5 (Efficiency):} Does the compact model achieve practical deployment efficiency?

\begin{table*}[t]
    \centering
    \small
    \setlength{\tabcolsep}{4pt}
    \renewcommand{\arraystretch}{0.9}
    \caption{Module-removal ablation. The full model corresponds to the SWD distilled student with all reported modules enabled.}
    \vspace{-1em}
    \label{tab:module_ablation}
    \resizebox{\textwidth}{!}{
    \begin{tabular}{l|cccc|cccc|c| c}
        \toprule
        & \multicolumn{4}{c|}{Basic Emotion} & \multicolumn{4}{c|}{Sentiment} & \multicolumn{1}{c|}{Fine-grained} &  \\
        Variant & MER2023 & MER2024 & MELD & IEMOCAP & MOSI & MOSEI & SIMS & SIMS v2 & OV-MERD+ & Mean \\
        \midrule
        Student + SWD-H + M-GRPO (Ours)           & \underline{75.66} & \underline{80.04} & \underline{57.02} & \underline{61.46} & \textbf{80.04} & \textbf{77.23} & \textbf{89.05} & \textbf{87.74} & \textbf{63.28} & \textbf{74.61} \\
        w/o M-GRPO & \textbf{77.40} & \textbf{80.69} & \textbf{58.13} & \textbf{65.60} & \underline{76.14} & \underline{76.75} &\underline{ 86.52} & \underline{84.66} & \underline{61.59} & \underline{74.16} \\
        w/o SWD-H  & 70.67 & 76.78 & 54.66 & 59.87 & 72.48 & 73.33 & 85.00 & 84.29 & 58.37 & 70.61 \\
        \bottomrule
    \end{tabular}}
    \vspace{-1em}
\end{table*}

\begin{table*}[t!]
    \centering
    \small
    \setlength{\tabcolsep}{4pt}
    \renewcommand{\arraystretch}{0.9}
    \caption{Comparison of distillation variants. Among the completed runs, SWD-H achieves the best overall mean score. }
    \vspace{-1em}
    \label{tab:distill_ablation}
    \resizebox{\textwidth}{!}{
    \begin{tabular}{lccccccccc c}
        \toprule
        Method & MER2023 & MER2024 & MELD & IEMOCAP & MOSI & MOSEI & SIMS & SIMS v2 & OV-MERD+ & Mean \\
        \midrule
        CE only & 70.67 & 76.78 & 54.66 & 59.87 & 72.48 & 73.33 & 85.00 & 84.29 & 58.37 & 70.61 \\
        KL on logits only \cite{KL2014}& 75.82 & 78.67 & \underline{57.55} & \underline{65.32} & 75.60 & 76.57 & 85.91 & 84.65 & 61.28 & 73.49 \\
        \midrule
        Sinkhorn Optimal Transport (Logits) \cite{NIPS2013_af21d0c9} & 74.33 & 80.03 & 57.18 & 61.94 & 75.72 & 76.78 & 86.14 & 84.62 & 59.68 & 72.94 \\
        Sinkhorn Optimal Transport (Hidden) \cite{NIPS2013_af21d0c9} & \underline{76.44} & 79.57 & 56.80 & 62.58 & \textbf{76.99} & \underline{76.79} & 86.41 & \textbf{85.45} & \textbf{61.98} & \underline{73.67} \\
        Position-aware Optimal Transport \cite{PAOT} & 75.46 & 78.91 & 55.60 & 61.12 & 76.04 & 75.19 & 85.95 & 84.75 & 60.61 & 72.63 \\
        Unbalanced Optimal Transport \cite{UnbalancedOT} & 75.58 & \underline{80.05} & 57.19 & 64.53 & \underline{76.21} & 76.82 & \textbf{87.26} & 83.99 & 61.03 & 73.63 \\
        SWD (Logits) \cite{kolouri2019} & 74.26 & 78.86 & 56.22 & 62.40 & 75.76 & \textbf{78.91} & \underline{86.75} & \underline{84.76} & 61.27 & 73.24\\
        SWD-H (Ours)& \textbf{77.40} & \textbf{80.69} & \textbf{58.13} & \textbf{65.60} & 76.14 & 76.75 & 86.52 & 84.66 & \underline{61.59} & \textbf{74.16}\\
        \bottomrule
    \end{tabular}}
    \vspace{-1em}
\end{table*}

\subsection{Main Results (RQ1)}

In the audio+video+text setting, Light-MER achieves a mean of 74.61, surpassing the 8B teacher at 73.93 by 0.68 points and the original AffectGPT at 69.77 by 4.84 points, all with a sub-1B student of 854M parameters (Table~\ref{tab:main_results}). Our 8B teacher upgrades the AffectGPT backbone from Qwen2.5-7B to Qwen3-8B; the original 7B AffectGPT is listed separately. Light-MER leads on 7 of 9 benchmarks, spanning basic emotion, sentiment, and fine-grained recognition.

Light-MER generalizes across modality subsets. Audio+text mean reaches 66.40 vs.\ teacher 65.07, and video+text mean reaches 65.71 vs.\ teacher 64.44. Across settings, the sub-1B student surpasses the 8B teacher on average while using $11\times$ fewer parameters (Table~\ref{tab:inference-profile}).

The student falls short of the teacher on MER2023 and CMU-MOSEI, both by roughly one point. On the remaining seven benchmarks, the distillation pipeline transfers multimodal patterns that compensate for the capacity gap (Section~\ref{sec:ablation}).

\textbf{(Answer to RQ1)}: A sub-1B student can match and surpass an 8B teacher in mean performance across diverse MER benchmarks when the distillation objective preserves the teacher's multimodal representation structure.

\subsection{Ablation Study (RQ2)}
\label{sec:ablation}

Progressive addition on Table~\ref{tab:module_ablation} shows the contribution of each component. The CE-only student obtains a mean of 70.61, already 95.5\% of the teacher's 73.93. Adding SWD-H raises the mean to 74.16, a 3.55-point gain that accounts for 88.8\% of the total improvement to the full model at 74.61. Adding M-GRPO contributes a further 0.45 points.

SWD-H concentrates its gains on basic emotion tasks, with MER2023 improving by 6.73 points and IEMOCAP by 5.73, while sentiment benchmarks show smaller gains.  SWD-H aligns hidden-state distributions (Figure~\ref{fig:hidden_logit_analysis}d vs.\ panel~b) with the teacher and provides the primary accuracy gain. 
Furthermore, M-GRPO refines the student's generation behavior, producing concise and interpretable outputs suitable for deployment  (Section~\ref{sec:efficiency}). A slight drop in HIT on the basic emotion benchmarks is expected, since our reward design encourages shorter, more explanation-focused emotion descriptions to improve inference efficiency. Compared with the original verbose outputs, such concise descriptions are less likely to explicitly cover benchmark label words, which can slightly reduce HIT. The clear gains on WAF and $F_s$ confirm the overall effectiveness of the proposed design.

\textbf{(Answer to RQ2):} Both components are necessary. SWD-H provides the dominant accuracy gain through hidden-state alignment. M-GRPO complements SWD-H by refining generation quality and improving inference efficiency, producing shorter and more interpretable emotion descriptions for deployment.

\begin{figure}[t!]
    \centering
    \includegraphics[width=0.42\textwidth]{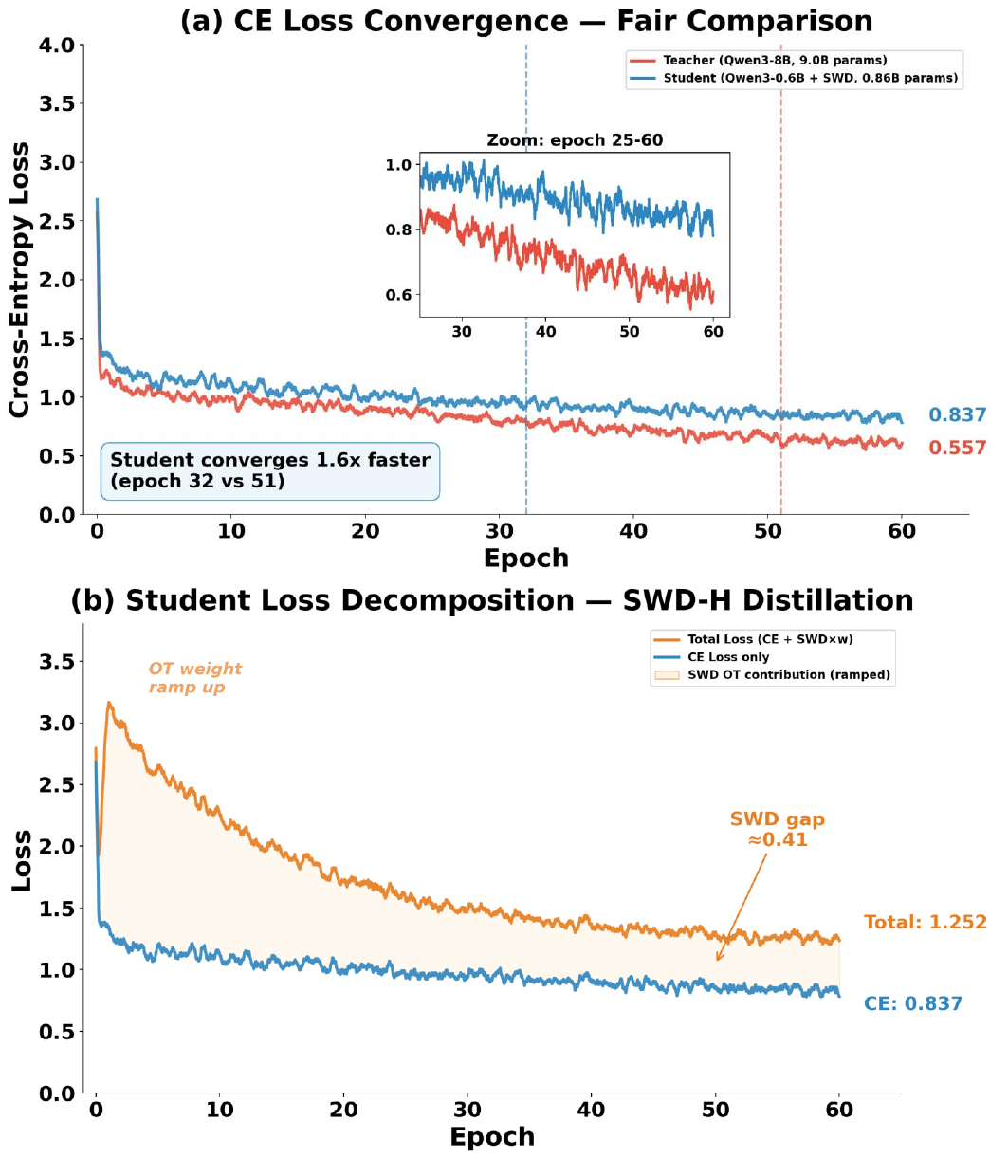}
    \vspace{-1.3em}
    \caption{Loss convergence and decomposition for SWD-H distillation. (a) Under a fair comparison, the student achieves smoother cross-entropy convergence and reaches the target loss about 1.6$\times$ earlier than the teacher reference, with stable late-stage optimization. (b) After OT weight ramp-up, the SWD-H term provides consistent geometric regularization while the total loss continues to decrease.}
    \label{fig:loss_convergence_detailed1}
    \vspace{-1.8em}
\end{figure}

\subsection{Distillation Variant Comparison (RQ3)}

Among eight distillation strategies (Table~\ref{tab:distill_ablation}), SWD-H achieves the highest overall mean of 74.16. All hidden-state OT methods except position-aware OT surpass KL-on-logits, which yields 73.49.

Two controlled comparisons isolate the source of improvement. Holding the loss function constant and switching from logits to hidden states improves Sinkhorn OT by 0.73 points, because hidden states encode how multimodal evidence is organized before decoding and carry richer distributional information than peaked output distributions (Section~\ref{subsec:observation}). Holding the alignment target at hidden states and switching from Sinkhorn to SWD adds another 0.49 points. SWD-H aligns teacher and student hidden states at answer-token positions. Each token contributes one hidden vector, so an answer of $L$ tokens gives only $L$ points to align. In MER, $L$ is typically 50--100, far fewer than the hidden dimension ($d=1024$). SWD projects these points onto random 1D directions and computes exact Wasserstein distances by sorting \cite{kolouri2019}, which works reliably even when $L$ is small. Sinkhorn OT \cite{NIPS2013_af21d0c9} solves an entropy-regularized transport plan iteratively, and the regularization strength must be set relative to $L$. With only 50--100 points, the plan becomes sensitive to this choice and the iterative updates can accumulate error across training steps. The convergence curves in Figure~\ref{fig:loss_convergence_detailed1} are consistent with this analysis, showing smoother convergence for SWD and oscillations for Sinkhorn.

The remaining OT variants each restrict how tokens are matched. Position-aware OT pairs each student token with the teacher token at the same sequence position, regardless of content, and yields the lowest mean among all OT variants. Unbalanced OT allows teacher tokens that have no similar student token to be left unmatched, and performs comparably to Sinkhorn. SWD on logits confirms the same pattern seen with Sinkhorn, with a similar margin of improvement when switching from logits to hidden states.

\textbf{(Answer to RQ3):} Two controlled comparisons identify the sources of SWD-H's advantage. Hidden-state alignment captures richer multimodal structure than logit-level matching. At the hidden-state level, SWD's closed-form sorting provides stable optimization on the short answer sequences typical of MER.

\begin{figure}[t!]
    \centering
     \includegraphics[width=\linewidth]{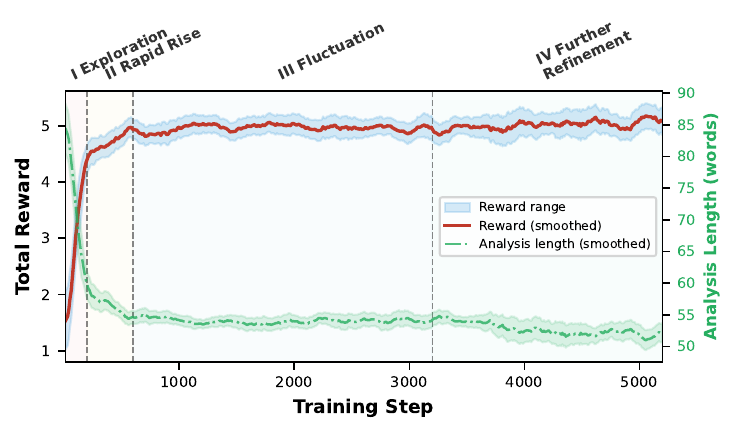}
     \vspace{-2.5em}
    \caption{Training dynamics of multi-reward GRPO optimization across four phases: exploration, rapid rise, fluctuation, and further refinement. The smoothed total reward increases steadily and stabilizes after approximately 500 steps, while the smoothed analysis length gradually decreases and converges to around 53 words by step 3200.}
    \label{fig:grpo_reward_phases}
    \vspace{-1.2em}
\end{figure}

\begin{table}[t!]
\centering
\renewcommand{\arraystretch}{0.9}
\caption{Description quality evaluated by GPT-5.4. We compare the student model before and after M-GRPO refinement across six dimensions.}
\vspace{-1em}
\label{tab:grpo-description-quality}
\begin{tabular}{lcccccc}
\toprule
\multirow{2}{*}{Stage} & \multicolumn{6}{c}{GPT-5.4~\cite{openai2026gpt54api}} \\
\cmidrule(lr){2-7}
 & Cls & Det & Flu & Sem & Con & Win\% \\
\midrule
Before M-GRPO   & 4.47 & 4.45 & 3.72 & 3.98 & 2.87 & 46 \\
After M-GRPO    & 4.89 & 3.62 & 4.24 & 3.83 & 4.07 & 54 \\
\bottomrule
\end{tabular}
\vspace{-1.2em}
\end{table}

\subsection{M-GRPO Analysis (RQ4)}
\label{sec:mgrpo}

Adding M-GRPO refinement to the SWD-H-distilled student raises the mean from 74.16 to 74.61. As noted in Section~\ref{sec:ablation}, the gain concentrates on sentiment benchmarks while emotion benchmarks show trade-offs. The brevity- and format-oriented reward shortens the reasoning trace, which leaves fewer tokens for enumerating fine-grained emotion cues. Multi-class emotion recognition relies more on such enumeration than binary sentiment polarity, consistent with the observed accuracy shift.

GPT-5.4 evaluation of description quality (Table~\ref{tab:grpo-description-quality}) shows improvements on classification accuracy (Cls), fluency (Flu), and conciseness (Con), all scored on a five-point Likert-type scale \cite{Likert}, with conciseness showing the largest single gain (+1.20). Detail coverage (Det), scored on the same scale, drops, consistent with the shorter output, while semantic (Sem) coherence remains stable, suggesting that the core emotion reasoning is preserved as surface detail is compressed. In head-to-head comparisons, the M-GRPO-refined model wins 54\% of samples.

Optimizing multiple reward components simultaneously risks one reward dominating at the expense of others. The training dynamics in Figure~\ref{fig:grpo_reward_phases} show that both total reward and analysis length stabilize after roughly 50 steps without such collapse, with analysis length settling near 53 words by step 3200. 

\textbf{(Answer to RQ4):} M-GRPO improves aggregate recognition accuracy (+0.45 mean), generation fluency, and conciseness at the cost of reduced detail coverage. For deployment, shorter and focused descriptions are preferable to exhaustive multimodal analyses.

\subsection{Inference Efficiency (RQ5)}
\label{sec:efficiency}

Light-MER supports two inference modes. In \emph{direct} mode, the model outputs label tokens without intermediate reasoning. In \emph{descriptive} mode, the model generates an emotion description, from which a label extractor derives the final prediction. Table~\ref{tab:inference-profile} profiles model size, memory footprint, and per-sample latency for both modes.

The sub-1B student achieves $11.0\times$ compression in parameters and FLOPs, and $7.9\times$ memory reduction from 20.04\,GB to 2.54\,GB. In descriptive mode, M-GRPO refinement reduces the average output from 110.5 to 70.8 words and latency from 4.6\,s to 3.1\,s, a 32.8\% reduction over the SWD-H-only student. The latency gain comes entirely from learned brevity (Section~\ref{sec:mgrpo}), not architectural modification. In direct mode, the M-GRPO student runs at 0.5\,s per sample. Direct mode eliminates the reasoning trace, reducing latency at the cost of the explicit evidence chain needed for open-set label extraction. For closed-set tasks where the label space is fixed, direct mode offers a viable low-latency alternative. 

\textbf{(Answer to RQ5):} The Light-MER student fits within 2.54\,GB memory and processes samples in 0.5\,s in direct mode or 3.1\,s in descriptive mode, with $11\times$ fewer parameters than the teacher. The profile shows that competitive generative MER is achievable with sub-1B models under the distillation-refinement pipeline.

\section{Conclusion}
In this study, we revisited generative multimodal emotion recognition from an efficiency perspective and asked whether high-quality MER truly requires deployment models larger than 1B parameters. We have shown  that it is not necessary. We presented Light-MER, a lightweight teacher-student framework that transfers multimodal emotion reasoning from a large MLLM to a sub-1B deployment model through geometry-aware hidden-state distillation and post-distillation policy refinement. Specifically, SWD-H preserves the latent structure of multimodal representations during compression, while M-GRPO further improves generation quality and efficiency with multi-reward optimization. Extensive experiments on nine benchmarks show that Light-MER not only matches but surpasses an 8B teacher in mean performance, while significantly reducing memory usage and inference latency. These results suggest that efficient generative MER does not require large deployment models, and that preserving representation geometry is a promising direction for compact multimodal reasoning systems.

\section{Acknowledgments}
Xuri Ge's research was (partially) supported by the Natural Science Foundation of Shandong Province with grant No. ZR2026QC1054.
\balance
\bibliographystyle{ACM-Reference-Format}
\bibliography{sample-base}
\clearpage

\appendix
\balance

\section{Further Details of Alternative OT Variants}

To ensure a controlled comparison in Table~4, all OT-based methods use the same shifted answer-token mask \(M\) and the same loss-weight warm-up schedule, while their projection spaces follow the corresponding implementations. For SWD-H, the projected hidden states in Eq.~(4) of the main paper are
\begin{equation}
\tilde{H}^{T}=H^{T}W_{T}, \qquad
\tilde{H}^{S}=\mathrm{LN}(H^{S}W_{S}),
\end{equation}
where \(W_{T}\in\mathbb{R}^{4096\times1024}\) is frozen and orthogonally initialized, and \(W_{S}\in\mathbb{R}^{1024\times1024}\) is trainable and identity-initialized.

For the coupling-based hidden-state baselines, the teacher and student states are projected into a common 256-dimensional space:
\begin{equation}
\bar{H}^{T}=\mathrm{LN}(H^{T}U_{T}), \qquad
\bar{H}^{S}=\mathrm{LN}(H^{S}U_{S}),
\end{equation}
where \(U_{T}\in\mathbb{R}^{4096\times256}\) is frozen and \(U_{S}\in\mathbb{R}^{1024\times256}\) is trainable. Using \(M_i=1\) iff \(y_{i+1}\neq-100\), we retain only supervised answer-token positions and obtain \(X=\{x_i\}_{i=1}^{n}\) and \(Y=\{y_j\}_{j=1}^{n}\), where \(x_i,y_j\in\mathbb{R}^{256}\) and \(n=\sum_i M_i\). The pairwise feature cost is cosine distance:
\begin{equation}
C_{ij}
=
1-
\frac{\langle x_i,y_j\rangle}
{\lVert x_i\rVert_2\lVert y_j\rVert_2}.
\end{equation}
We use uniform marginals \(\mu=\nu=\frac{1}{n}\mathbf{1}_n\) and the balanced transport polytope
\begin{equation}
\mathcal{U}(\mu,\nu)
=
\left\{
P\in\mathbb{R}_{+}^{n\times n}
\mid
P\mathbf{1}_n=\mu,\;
P^{\top}\mathbf{1}_n=\nu
\right\}.
\end{equation}
Each OT loss replaces \(L_{\mathrm{SWD}}\) in Eq.~(7) and is combined with \(L_{\mathrm{CE}}\) using the same warm-up schedule. For logit-level variants, teacher and student logits at supervised positions are used directly because they share the same vocabulary space.
Unless otherwise stated, each OT loss below replaces \(L_{\mathrm{SWD}}\) in Eq.~(7) of the main paper and is combined with \(L_{\mathrm{CE}}\) using the same warm-up schedule. For the logit-level variants reported in Table~4, we use the same formulations after replacing \(h^{T}_{i}\) and \(h^{S}_{j}\) with teacher and student logits at the supervised positions.

\subsection{Sinkhorn Optimal Transport}

Sinkhorn OT~\cite{NIPS2013_af21d0c9} solves an entropically regularized transport problem:
\begin{equation}
L_{\mathrm{Sinkhorn}}
=
\min_{P \in \mathcal{U}(\mu,\nu)}
\langle P, C \rangle
+
\varepsilon
\sum_{i=1}^{n}\sum_{j=1}^{n}
P_{ij}\bigl(\log P_{ij}-1\bigr),
\end{equation}
where \(\varepsilon > 0\) is the entropy regularization coefficient and
\begin{equation}
\langle P, C \rangle
=
\sum_{i=1}^{n}\sum_{j=1}^{n} P_{ij} C_{ij}.
\end{equation}
Let
\begin{equation}
K = \exp(-C/\varepsilon).
\end{equation}
The transport plan is then obtained by Sinkhorn-Knopp matrix scaling:
\begin{equation}
u^{(k+1)} = \mu \oslash \bigl(Kv^{(k)}\bigr), \qquad
v^{(k+1)} = \nu \oslash \bigl(K^{\top}u^{(k+1)}\bigr),
\end{equation}
which yields
\begin{equation}
P^{(k+1)}
=
\mathrm{Diag}\!\bigl(u^{(k+1)}\bigr)\,
K\,
\mathrm{Diag}\!\bigl(v^{(k+1)}\bigr).
\end{equation}
After \(K_{\mathrm{iter}}\) iterations, the loss is evaluated as
\begin{equation}
L_{\mathrm{Sinkhorn}}
=
\left\langle P^{(K_{\mathrm{iter}})}, C \right\rangle.
\end{equation}
Compared with SWD-H, Sinkhorn OT explicitly optimizes a dense \(n \times n\) coupling matrix and thus incurs iterative computation with complexity dominated by repeated matrix scaling and transport-plan updates.

\subsection{Position-aware Optimal Transport}

The original position-aware OT idea~\cite{PAOT} introduces an additional structural or positional bias into the transport plan. Since the original formulation is designed for network alignment, in our autoregressive hidden-state setting we retain its core intuition and adapt it to token sequences by penalizing transport between distant decoding positions. Specifically, we define a position cost matrix
\begin{equation}
G_{ij}
=
\left(
\frac{i-j}{\max(n-1,1)}
\right)^{2},
\end{equation}
and optimize
\begin{equation}
L_{\mathrm{PAOT}}
=
\min_{P \in \mathcal{U}(\mu,\nu)}
\langle P, C \rangle
+
\lambda_{\mathrm{pos}} \langle P, G \rangle
+
\varepsilon
\sum_{i=1}^{n}\sum_{j=1}^{n}
P_{ij}\bigl(\log P_{ij}-1\bigr),
\end{equation}
where \(\lambda_{\mathrm{pos}} > 0\) controls the strength of the positional bias. Here,
\begin{equation}
\langle P, G \rangle
=
\sum_{i=1}^{n}\sum_{j=1}^{n} P_{ij} G_{ij}.
\end{equation}
When \(\lambda_{\mathrm{pos}}\) is large, the optimal coupling is biased toward the diagonal, encouraging each student token to align with teacher tokens at similar sequence positions. This positional prior can be beneficial when teacher and student preserve nearly identical token-level ordering, but it can also suppress content-based matching when semantically corresponding hidden states shift across positions after compression.

\subsection{Unbalanced Optimal Transport}

Unbalanced OT~\cite{UnbalancedOT} relaxes the hard marginal constraints and allows part of the mass to remain unmatched. Instead of enforcing \(P \in \mathcal{U}(\mu,\nu)\), it penalizes deviations from the desired marginals:
\begin{equation}
\begin{aligned}
L_{\mathrm{UOT}}
=
\min_{P \ge 0}\;&
\langle P, C \rangle
+
\tau\, \mathrm{KL}\!\left(P\mathbf{1}_{n}\,\middle\|\,\mu\right)
+
\tau\, \mathrm{KL}\!\left(P^{\top}\mathbf{1}_{n}\,\middle\|\,\nu\right) \\
&+
\varepsilon
\sum_{i=1}^{n}\sum_{j=1}^{n}
P_{ij}\bigl(\log P_{ij}-1\bigr).
\end{aligned}
\end{equation}
where \(\tau > 0\) is the marginal-relaxation coefficient. We use the generalized KL divergence
\begin{equation}
\mathrm{KL}(a \| b)
=
\sum_{i=1}^{n}
a_i \log \frac{a_i}{b_i} - a_i + b_i.
\end{equation}
As \(\tau \rightarrow \infty\), this formulation reduces to the balanced OT case. Unbalanced OT is more tolerant to missing or extra modes because low-confidence mass can be discarded instead of being forcibly transported. However, it introduces an additional balancing hyperparameter and still requires iterative optimization over a full transport plan.

Overall, all three alternatives above require optimizing an explicit \(n \times n\) coupling matrix. By contrast, SWD-H in the main paper avoids full transport-plan estimation by projecting hidden states onto random one-dimensional directions and computing exact 1D OT through sorting, which is the main reason for its lower overhead in repeated hidden-state matching during distillation.

\section{Implementation Details}
Table~\ref{tab:swd_grpo_hyperparams} summarizes the main hyperparameter settings used in our two-stage training pipeline. In the first stage, SWD-H distillation aligns the student with the teacher at the hidden-state level, using 200 SWD projections, a Wasserstein order of 1, and a gradually increased distillation weight. In the second stage, GRPO further optimizes the policy with multiple sampled completions per prompt, controlled generation settings, and a standard PPO-style clipping strategy. Overall, these settings provide a representative configuration for both stable distillation and subsequent policy fine-tuning.
\begin{table}[t]
\centering
\caption{Hyperparameters for SWD-H distillation and GRPO fine-tuning (representative settings).}
\renewcommand{\arraystretch}{1.2}
\label{tab:swd_grpo_hyperparams}
\begin{tabular}{@{}lccc@{}}
\toprule
\multicolumn{4}{c}{\textbf{SWD-H Distillation (hidden-state alignment)}} \\ \midrule
\textbf{Parameter} & \multicolumn{3}{c}{\textbf{Value}} \\ \midrule
SWD projections $K$ & \multicolumn{3}{c}{200} \\
SWD order $p$ (Wasserstein) & \multicolumn{3}{c}{1} \\
Distillation weight (ramped) & \multicolumn{3}{c}{1.0} \\
Ramp steps for distillation weight & \multicolumn{3}{c}{5000} \\
Max sequence length & \multicolumn{3}{c}{1024} \\
Training epochs & \multicolumn{3}{c}{60} \\
Iterations per epoch & \multicolumn{3}{c}{5000} \\
Optimizer LR (init / min) & \multicolumn{3}{c}{$5{\times}10^{-5}$ / $5{\times}10^{-6}$} \\
Warmup steps & \multicolumn{3}{c}{5000} \\
Weight decay & \multicolumn{3}{c}{0.01} \\
LR schedule & \multicolumn{3}{c}{linear warmup + cosine} \\
Batch size $\times$ grad accum. & \multicolumn{3}{c}{$3 \times 8$} \\
Mixed precision & \multicolumn{3}{c}{enabled} \\
\midrule
\multicolumn{4}{c}{\textbf{GRPO (policy optimization after SWD-H)}} \\ \midrule
Completions per prompt $K$ & \multicolumn{3}{c}{4} \\
Max new tokens (generation) & \multicolumn{3}{c}{120} \\
Sampling temperature / top-$p$ & \multicolumn{3}{c}{0.9 / 0.9} \\
PPO-style clip $\varepsilon$ & \multicolumn{3}{c}{0.1} \\
KL coefficient (to reference) & \multicolumn{3}{c}{1.0} \\
Policy LR & \multicolumn{3}{c}{$1{\times}10^{-7}$} \\
Gradient accumulation & \multicolumn{3}{c}{4} \\
\bottomrule
\end{tabular}
\end{table}

\section{Additional Evaluation Results}

\begin{table*}[t]
    \centering
    \small
    \setlength{\tabcolsep}{4pt}
    \renewcommand{\arraystretch}{0.9}
    \caption{Main results on the evaluation suite. Primary metrics are reported for each dataset: ACC for sentiment. AffectGPT baseline numbers are reproduced using the official open-source code.}
    \vspace{-1em}
    \label{tab:main_results}
    \begin{tabular}{lr|ccc|cccc}
        \toprule
        \multirow{2}{*}{Model} & \multirow{2}{*}{LLM Param} & \multicolumn{3}{c|}{Modality} & \multicolumn{4}{c}{Sentiment (ACC)} \\
        & & A & V & T & MOSI & MOSEI & SIMS & SIMS v2  \\
        \midrule
        \multicolumn{9}{l}{\textbf{Audio+Text MLLMs}} \\
        OneLLM \cite{OneLLM}       & 7B & $\checkmark$ & $\times$     & $\checkmark$ & 64.48 & 54.18 &  63.92 &  62.46 \\
        SECap \cite{SECap}        & 7B & $\checkmark$ & $\times$     & $\checkmark$ & 56.71 & 53.85 & 62.89 &  60.92 \\
        PandaGPT \cite{su2023}     & 7B & $\checkmark$ & $\times$     & $\checkmark$ & 65.85 & 60.73 & 62.37 & 58.84 \\
        Qwen-Audio \cite{Qwen-Audio}  & 7B & $\checkmark$ & $\times$     & $\checkmark$ & 71.49 & 51.16 & 73.45 & 68.17 \\
        SALMONN \cite{SALMONN}     & 13B & $\checkmark$ & $\times$     & $\checkmark$ & 81.25 & 66.90 &  69.85 & 67.07 \\
        \midrule
        \multicolumn{9}{l}{\textbf{Video+Text MLLMs}} \\
        Otter \cite{Otter}           & 9B & $\times$     & $\checkmark$ & $\checkmark$  & 54.27 & 50.77 & 60.57 & 56.20 \\
        Video-LLaVA  \cite{video-llava}    & 7B & $\times$     & $\checkmark$ & $\checkmark$  & 57.62 & 64.20 & 54.64 & 59.28  \\
        PandaGPT \cite{su2023}        & 7B & $\times$     & $\checkmark$ & $\checkmark$  & 60.21 & 65.55 &  61.60 & 65.31  \\
        Video-ChatGPT \cite{video-chatgpt}   & 7B & $\times$     & $\checkmark$ & $\checkmark$ & 57.77 & 65.66 &  64.43 &  66.85 \\
        LLaMA-VID  \cite{llamavid}      & 7B & $\times$     & $\checkmark$ & $\checkmark$ &  62.65 & 66.21 & 68.81 & 67.73 \\
        VideoChat  \cite{videochat}      & 7B & $\times$     & $\checkmark$ & $\checkmark$ & 65.09 &  63.02 &  69.33 & 72.12 \\
        Chat-UniVi \cite{chatunivi}      & 7B & $\times$     & $\checkmark$ & $\checkmark$ &  57.62 &  67.47 &  67.78 &  67.18 \\
        mPLUG-Owl  \cite{mPLUG}      & 7B & $\times$     & $\checkmark$ & $\checkmark$ & 72.26 & 73.17 & 71.65 & 74.97 \\
        \midrule
        \multicolumn{9}{l}{\textbf{Audio+Video+Text MLLMs}} \\
        PandaGPT  \cite{su2023}       & 7B & $\checkmark$ & $\checkmark$ & $\checkmark$  &  62.80 & 68.82 & 67.78 & 67.40 \\
        Emotion-LLaMA \cite{Emotion-LLaMA}   & 7B & $\checkmark$ & $\checkmark$ & $\checkmark$  & 66.31 & 67.25 & 78.61 & 77.39 \\
        AffectGPT \cite{affectgpt} & 7B & $\checkmark$ & $\checkmark$ & $\checkmark$ & \textbf{81.77} & \underline{77.86} & 81.98 & 80.01 \\
        AffectGPT(Qwen3-8B)& 8B & $\checkmark$ & $\checkmark$ & $\checkmark$ & 78.35 & \textbf{78.26} & \underline{88.14} & \underline{86.94} \\
        \textbf{Light-MER(Ours)}  & \textbf{0.6B} & $\checkmark$ & $\checkmark$ & $\checkmark$ & \underline{79.85} & 76.99 & \textbf{88.86} & \textbf{87.72} \\
        \bottomrule
    \end{tabular}
    \vspace{-0.5em}
\end{table*}

\begin{table*}[t]
    \centering
    \small
    \setlength{\tabcolsep}{4pt}
    \renewcommand{\arraystretch}{0.9}
    \caption{Supplementary comparison on the four basic-emotion benchmarks using the secondary \textit{mscore} metric. Since \textit{mscore} is reported in addition to the primary HIT metric, we compare only the most directly comparable generative MER baselines, i.e., AffectGPT and AffectGPT(Qwen3-8B). AffectGPT baseline numbers are reproduced using the official open-source code.}
    \vspace{-1em}
    \label{tab:main_results}
    \begin{tabular}{lr|ccc|cccc}
        \toprule
        \multirow{2}{*}{Model} & \multirow{2}{*}{LLM Param} & \multicolumn{3}{c|}{Modality} & \multicolumn{4}{c}{Basic Emotion (\textit{mscore})} \\
        & & A & V & T & MER2023 & MER2024 & MELD & IEMOCAP \\
        \midrule
        AffectGPT \cite{affectgpt} & 7B & $\checkmark$ & $\checkmark$ & $\checkmark$ & 46.08 & 46.78 & 36.17 & 46.11 \\
        AffectGPT(Qwen3-8B) & 8B & $\checkmark$ & $\checkmark$ & $\checkmark$ & 50.55 & 54.29 & 36.66 & 53.95 \\
        \textbf{Light-MER(Ours)} & \textbf{0.6B} & $\checkmark$ & $\checkmark$ & $\checkmark$ & \textbf{53.91} & \textbf{56.52} & \textbf{38.66} & \textbf{56.87} \\
        \bottomrule
    \end{tabular}
    \vspace{-0.5em}
\end{table*}

\subsection{Supplementary Analysis on \textit{mscore}}

For the basic-emotion benchmarks, our framework is optimized and selected using HIT rather than \textit{mscore}, following the standard evaluation protocol in which HIT serves as the primary metric and \textit{mscore} is reported only as a supplementary metric. In evaluation, free-form generated outputs are first mapped into the emotion-wheel-based candidate label space. The two metrics then capture different aspects of prediction behavior. HIT measures whether the mapped ground-truth basic emotion is retained in the mapped prediction set, while \textit{mscore} is more sensitive to how many plausible affective labels remain after mapping and is therefore relatively more precision-oriented.

Concretely, for a test set of size $N$, let $y_i$ denote the ground-truth basic emotion label for sample $i$, let $\hat{Y}_i$ denote the corresponding free-form generated output, and let $G_{w_k}(\cdot)$ denote the mapping function into the candidate label space. HIT is computed as
\begin{equation}
\mathrm{HIT}=\frac{1}{N}\sum_{i=1}^{N}\mathbf{1}\!\left[G_{w_k}(y_i)\in G_{w_k}(\hat{Y}_i)\right],
\end{equation}
where $\mathbf{1}[\cdot]$ is the indicator function. Therefore, HIT only checks whether the correct basic emotion is covered after mapping, whereas \textit{mscore} further penalizes predictions that retain multiple semantically related affect terms in the candidate label space.

This distinction is particularly important in our setting because Light-MER is a generative multimodal emotion model rather than a one-hot classifier. Our SWD-H objective is designed to preserve the teacher's latent multimodal geometry, while M-GRPO encourages accurate, concise, and well-formatted emotion-label responses. As a result, the model can produce one or more open-set emotion labels rather than being restricted to a single benchmark label. After these generated labels are mapped back into the benchmark label space, multiple semantically related affect terms may still be preserved, which can keep HIT stable while making \textit{mscore} more sensitive.

For this reason, in the supplementary \textit{mscore} analysis, we do not aim to compare against all models indiscriminately. Instead, we restrict the comparison to the most directly comparable generative MER baselines, namely AffectGPT and AffectGPT(Qwen3-8B), because they share the same generative multimodal reasoning paradigm and are therefore the most appropriate reference models under this metric. This choice is made to ensure a fair and meaningful comparison for generative reasoning-based MLLMs, rather than to exclude other baselines. As shown in Table~\ref{tab:main_results}, Light-MER achieves the best \textit{mscore} on all four basic-emotion benchmarks, reaching 53.91 on MER2023, 56.52 on MER2024, 38.66 on MELD, and 56.87 on IEMOCAP. Compared with AffectGPT(Qwen3-8B), Light-MER improves \textit{mscore} by +3.36, +2.23, +2.00, and +2.92 points, respectively. These results show that even under the stricter supplementary \textit{mscore} criterion, our distilled sub-1B model preserves and further improves the benchmark-compatible prediction precision within the same generative MER setting.

\subsection{Analysis of ACC on sentiment benchmarks}

For the sentiment benchmarks, the official protocol uses weighted average F-score as the primary metric because of label imbalance, while ACC is reported as a secondary metric. In the evaluation procedure, the generated open-vocabulary outputs are first mapped to $\{positive, negative, neutral\}$, after which neutral labels are removed and both weighted F-score and ACC are computed on the remaining non-neutral samples as a binary positive-versus-negative classification task. Therefore, ACC in this benchmark suite mainly measures whether the model preserves coarse sentiment polarity, rather than whether it fully captures sentiment intensity or subtle neutral-boundary ambiguity.

 In the full audio+video+text setting, Light-MER outperforms the 8B teacher on three of the four sentiment datasets (MOSI, SIMS, and SIMS v2), and is only slightly below the teacher on MOSEI; correspondingly, the four-dataset average ACC is 83.36 for Light-MER versus 82.92 for the teacher. This pattern is consistent with the design of our method. SWD-H transfers the teacher’s multimodal latent structure and preserves the cues necessary for sentiment polarity judgment, while M-GRPO suppresses unnecessary reasoning and encourages concise, well-formatted emotion-label outputs. Such behavior is beneficial when the downstream evaluator collapses free-form outputs into a coarse polarity label. The remaining gap on MOSEI is unsurprising, since MOSEI contains more diverse and ambiguous sentiment cases; once open-vocabulary descriptions are projected to $\{positive,negative,neutral\}$, small shifts around the polarity boundary can flip the final ACC outcome. Overall, the ACC results show that Light-MER preserves sentiment polarity very well, even though ACC is not the target used for checkpoint selection.

\section{LLM-based Evaluation}

\begin{figure*}[t]
    \centering
    \includegraphics[width=0.8\linewidth]{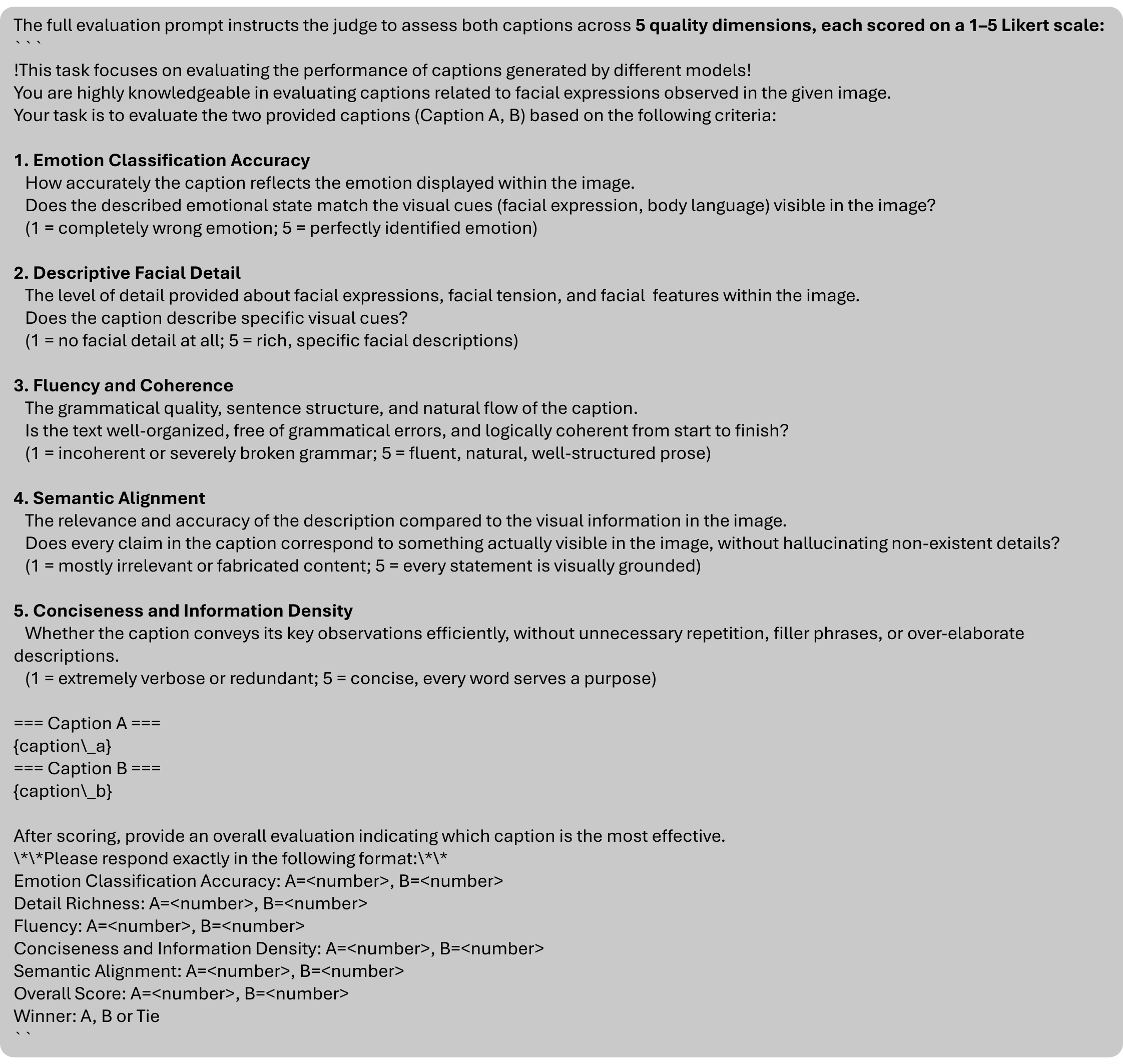}
    \caption{LLM-as-a-judge evaluation pipeline for emotion-caption quality. For each sampled test clip, we extract 8 representative frames and pair them with the corresponding pre-GRPO caption (Caption A) and post-GRPO caption (Caption B). GPT-5.4 receives the 8 frames and both captions in a single multimodal request, scores them on emotion classification accuracy, descriptive facial detail, fluency and coherence, semantic alignment, and conciseness/information density, and then returns per-dimension scores, overall scores, and a final winner decision.}
    \label{fig:llm_eval_pipeline}
\end{figure*}
\begin{figure*}[t]
    \centering
    \includegraphics[width=0.63\linewidth]{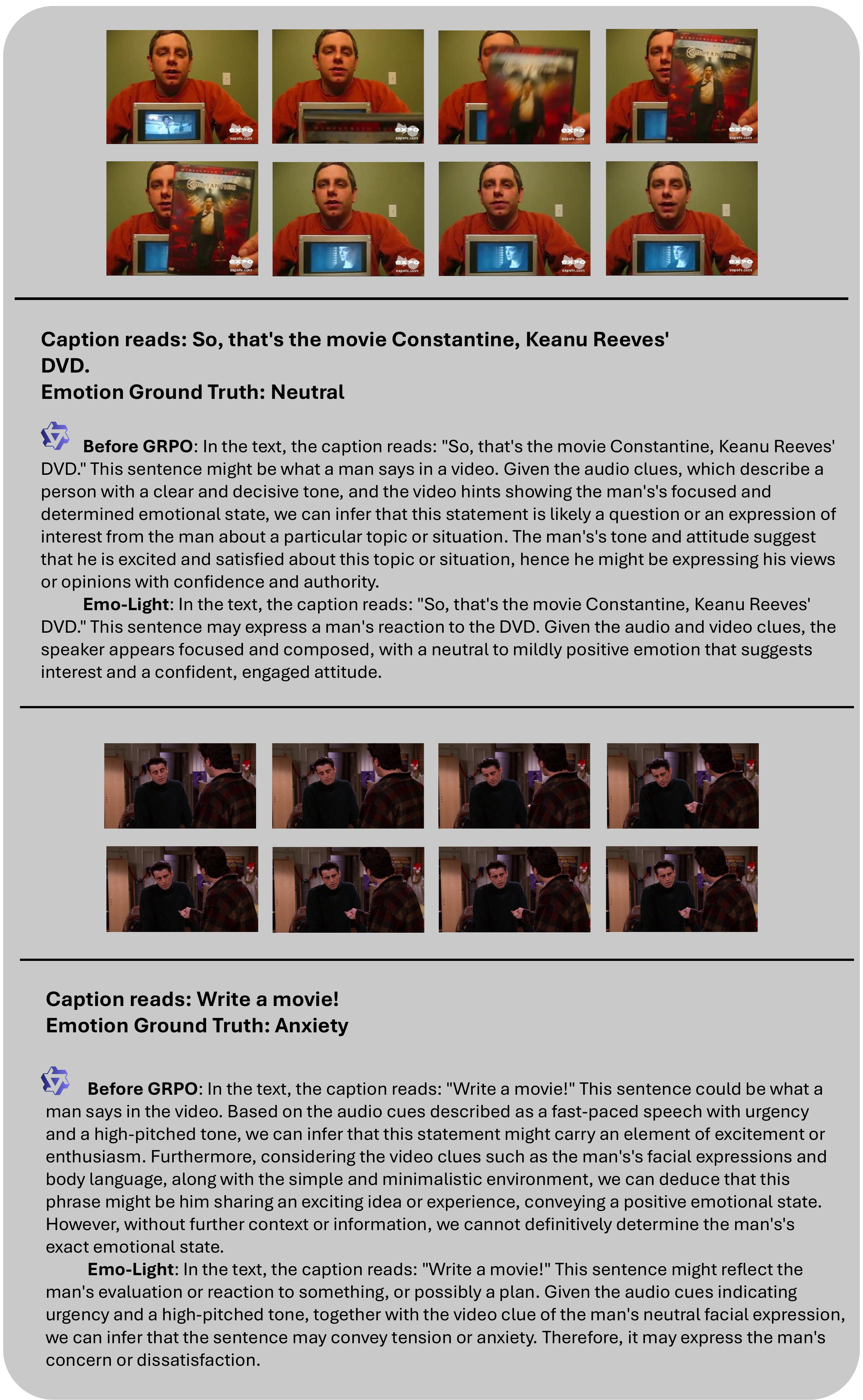}
    \caption{Qualitative examples of caption refinement before and after M-GRPO. Each case shows eight sampled frames from the clip, the subtitle, and the ground-truth emotion label. Compared with the pre-GRPO output, the post-GRPO student (shown as \textit{Emo-Light} in the figure) produces shorter and more grounded descriptions. In the top \textit{Neutral} example, the refined model removes unsupported claims about decisiveness and confidence and backs off to a more neutral interpretation. In the bottom \textit{Anxiety} example, it shifts from a speculative excitement-oriented reading to a tension/anxiety-oriented interpretation that is closer to the ground truth.}
    \label{fig:qualitative_mgrpo}
\end{figure*}

To complement task-level benchmark metrics, we further conduct an LLM-as-a-judge evaluation of caption quality before and after M-GRPO refinement. In this analysis, Caption A denotes the student model after SWD-H distillation but before GRPO refinement, while Caption B denotes the same student after M-GRPO refinement. The A/B assignment is randomized and is randomized across samples.

The LLM-based evaluation is performed on a balanced subset of \(10{,}000\) test samples drawn from seven multimodal emotion benchmarks: MER2023, MER2024, MELD, IEMOCAP, CMU-MOSI, CMU-MOSEI, and SIMS. These datasets cover both categorical emotion classification and sentiment regression settings, and include both Chinese and English samples.

\subsection{Judge Model and Input Construction}

We use GPT-5.4 as the judge model. Each evaluation request is multimodal and contains two parts: (1) the extracted eight frames, encoded as a base64 JPEG and passed through the \texttt{image\_url} field; and (2) a structured prompt containing Caption A, Caption B, and detailed scoring rubrics.

The judge scores both captions on five dimensions using a 1--5 Likert scale:
\begin{enumerate}
    \item \textbf{Emotion Classification Accuracy}: whether the caption correctly reflects the emotion visible in the image;
    \item \textbf{Descriptive Facial Detail}: the richness of facial-expression descriptions;
    \item \textbf{Fluency and Coherence}: grammatical quality and naturalness of the text;
    \item \textbf{Semantic Alignment}: whether the content is visually grounded and free of hallucinated details;
    \item \textbf{Conciseness and Information Density}: whether the caption expresses its observations efficiently without redundancy.
\end{enumerate}
The overall score is defined as the sum of these five dimensions, and the judge additionally returns a final winner decision from \(\{\mathrm{A}, \mathrm{B}, \mathrm{Tie}\}\).

\subsection{Aggregate Metrics and Findings}

From the parsed evaluation results, we compute the mean score of each dimension for Caption A and Caption B, the winner distribution over \(\{\mathrm{A}, \mathrm{B}, \mathrm{Tie}\}\), and the word-count statistics before and after M-GRPO refinement.

As summarized in Table~5 of the main paper, M-GRPO improves the judged quality of the generated descriptions in several important dimensions. Emotion classification accuracy rises from \(4.47\) to \(4.89\), fluency improves from \(3.72\) to \(4.24\), and conciseness increases substantially from \(2.87\) to \(4.07\). At the same time, descriptive facial detail decreases from \(4.45\) to \(3.62\), while semantic alignment remains broadly stable (\(3.98\) vs.\ \(3.83\)). In head-to-head pairwise comparison, the post-GRPO model wins \(54\%\) of samples, compared with \(46\%\) for the pre-GRPO model.

These results are consistent with the design goal of M-GRPO. The reward function explicitly encourages emotion-label accuracy, canonical formatting, concise outputs, suppression of unnecessary reasoning, clean stopping, and an appropriate label count. The refined model therefore trades some surface-level detail for stronger conciseness and more focused outputs. This trade-off is desirable for deployment, where shorter and more focused responses are preferable to overly verbose descriptions.

\section{Visualization}

Figure~\ref{fig:qualitative_mgrpo} provides two representative examples comparing the student outputs before and after M-GRPO refinement. In our setup, the visual input to the LLM consists of eight sampled frames, which are jointly provided to the model during caption generation. Accordingly, this qualitative figure shows the same eight frames used as input, together with the subtitle. Unlike the LLM-as-a-judge pipeline, which evaluates each sample based on a single middle frame, the figure reflects the full visual context actually available to the model. 

In the first example, the ground-truth emotion is \textit{Neutral}. The pre-GRPO caption is long and over-interprets the input with unsupported claims such as a \emph{clear and decisive tone} and \emph{confidence and authority}. After M-GRPO, the model produces a much shorter description that stays closer to the observable evidence and backs off from strong personality-level inferences, yielding a more neutral reading.

In the second example, the ground-truth emotion is \textit{Anxiety}. Before refinement, the model gives a verbose and hesitant explanation that mixes urgency with \emph{excitement} and \emph{enthusiasm}, leading to a less accurate affective interpretation. After M-GRPO, the model focuses on tension, urgency, and concern, shifting the description toward \textit{anxiety}, which is more consistent with the benchmark label.

Overall, these qualitative cases clarify the role of M-GRPO: it does not merely shorten the output, but also suppresses speculative and weakly grounded statements, making the final description more concise, more focused, and more aligned with the target emotion. This qualitative trend is consistent with the GPT-5.4 evaluation in Table~5, where post-GRPO captions improve on emotion accuracy, fluency, and conciseness while using fewer words on average.

\balance

\end{document}